\documentclass{article}

\usepackage{arxiv}
\usepackage[utf8]{inputenc} % allow utf-8 input
\usepackage[T1]{fontenc}    % use 8-bit T1 fonts
\usepackage{hyperref}       % hyperlinks
\usepackage{url}            % simple URL typesetting
\usepackage{booktabs}       % professional-quality tables
\usepackage{amsfonts}       % blackboard math symbols
\usepackage{nicefrac}       % compact symbols for 1/2, etc.
\usepackage{microtype}      % microtypography
\usepackage{amsmath,amssymb}
\usepackage{cleveref}       % smart cross-referencing
\usepackage{lipsum}         % Can be removed after putting your text content
\usepackage{graphicx}
\usepackage{natbib}
\usepackage{doi}
\usepackage{listings}

\usepackage{bbold}

\renewcommand{\vec}[1]{\mathbf{#1}}

\begin{document}

\title{A Principled Method for the Creation of Synthetic Multi-fidelity Data Sets}

\author{Clyde Fare \\
IBM Research Europe - Daresbury\\ UK \\
\texttt{clyde.fare@ibm.com} \\
\And
Peter Fenner \\
IBM Research Europe - Daresbury\\ UK \\
\texttt{peter.fenner1@ibm.com} \\
\And
Edward Pyzer-Knapp \\
IBM Research Europe - Daresbury\\ UK \\
\texttt{epyzerk3@uk.ibm.com} \\
}

%\renewcommand{\shortauthors}{Fare, Fenner and Pyzer-Knapp}
\maketitle
\begin{abstract}
  Multifidelity and multioutput optimisation algorithms are of active interest in many areas of computational design as they allow cheaper computational proxies to be used intelligently to aid experimental searches for high performing species. Characterisation of these algorithms involves benchmarks that typically either use analytic functions or existing multifidelity datasets. However, analytic functions are often not representative of relevant problems, while preexisting datasets do not allow systematic investigation of the influence of characteristics of the lower fidelity proxies. To bridge this gap, we present a methodology for systematic generation of synthetic fidelities derived from preexisting datasets. This allows for construction of benchmarks that are both representative of practical optimisation problems while also allowing systematic investigation of the influence of the lower fidelity proxies.
\end{abstract}


\keywords{Multifidelity, Generative Modelling, Gaussian Process, Coregionalization, Bayesian Optimization}


\section{Introduction}
Building and deploying data-derived models has become a ubiquitous activity in many fields. Generation of vastly complex models is now possible where the limiting factor within many fields of application is a sufficiently large and diverse dataset with which one can train models. For many tasks, financial or time costs limit the collection of data at the desired accuracy, and so methods which are able to take advantage of multiple disparate sources of data are beginning to become popular. One example of this is the emerging area of multifidelity optimization \cite{Song2018, Huang2006, Kandasamy2017} where optimisation algorithms are able to make use of queries of approximate variants or lower 'fidelities' of the intended optimisation target. 

For single fidelity blackbox optimisation problems there are well known benchmark suites \cite{Hansen2021} that allow comparison of different algorithms. However preparation of such benchmarks for multifidelity optimisation is challenging due to the need not only to specify diverse and relevant optimisation problems but also multiple different proxies. Recent libraries of analytic functions suitable for multifidelity optimisation have been developed \cite{VanRijn2020, mainini2022} as have some general benchmark suites \cite{hwang2018, eggensperger2021}. However the lower fidelity approximations of current benchmarks do not offer fine grained tools for controlling the behaviour of the low fidelity proxies. This limits investigation of the behaviour of the optimization algorithms on the characteristics of the different fidelities and risks biasing development of novel multifidelity algorithms. 

To build better multifidelity tooling there is a  need to be able to systematically investigate how different algorithms perform on different problems where the number of quality of the lower fidelity proxies can explicitly varied. To further this goal we present a simple means to generate sets of synthetic fidelities based on any single or multifidelity dataset. These synthetic fidelities exhibit a controllable degree of correlation to the reference and known lower fidelity data and thus can be used to benchmark multifidelity optimisation algorithms in a systematic fashion. 

The creation of such synthetic fidelities falls under generative modelling of functions for which much prior work exists \cite{Blundell2015, Rasmussen2004, Gal2016, hernandez-lobatoc15, Harshvardhan2020} but to our knowledge there has been little focus on the generative modelling of proxy functions with specific consideration of the relation between the generated proxies and the known reference function(s) those proxies are based upon.

% \section{Terminology}

% For purposes of clarity we distinguish between three types of correlation or covariance. The task covariance, the posterior mean covariance and the sample covariance. 

% Sample covariance: we suppose that we have a set of points corresponding to some domain $x of X$ and we have evaluated our reference function at those points to give us $y0 of Y0$, we also have evaluated a synthetic proxy at the same points to give us $y1 of Y1$ the sample covariance is then is the covariance defined by $\sigma y0y1$.

% Posterior Mean covariance: we suppose we have used an MOGP with a coregionalisation kernel to calculate the posterior over a reference function and a synthetic fidelity. We again have some set of points corresponding to a domain $x of X$ and we determine the posterior predicted mean for the reference function $u0 of U0$ and for the synthetic fidelity $u1 of U1$. The posterior mean covariance is then defined by $\sigma u0u1$

% Task covariance: we suppose we have an MOGP with a coregionalisation kernel. The coregional kernel has the form: []. Where T_ij is a task covariance matrix. Where the element ij specifies the task covariance between task i and task j.

\section{Algorithm Outline}

Conceptually our algorithm involves:

\begin{itemize}
\item Taking preexisting single or multifidelity data along with a specification for the desired degree of correlation to that existing fidelity data. 
\item Fitting a Bayesian model which includes extraction of characteristics common to the data. 
\item Modifying the model to include a posterior over an additional synthetic fidelity (for which there is no data).
\item Controlling underlying parameters of the model such that a sample drawn from the posterior over the synthetic fidelity possesses the desired degrees of correlation to the preexisting data.
\end{itemize}

We choose a Gaussian Process (GP) \cite{Rasmussen2004} as the underlying methodology used to model function distributions as GPs provide an analytically tractable posterior formulation. In particular we make use of a Multiple Output Gaussian Process (MOGP) and as we are interested in systematic control of the posterior we choose to use a simple coregionalisation kernel architecture \cite{bonilla2008multi, Alvarez2012a} to model inter function covariances, which is paired with a typical GP kernel used to model intra function covariances.

The training regime we adopt is to first take any data associated with the reference ground truth function and any available lower fidelity proxies and use this data to fit the hyperparameters for the MOGP via maximisation of the log marginal likelihood of the data. As these hyperparameters include the kernel used to model the intra-function variances this process captures characteristics (lengthscales, spectral mixtures, etc.) common to the reference function and any given proxy functions. As a consequence samples drawn from the GP prior using these optimised hyperparameters result in functions that share these characteristic. 

Having constructed a model of the available data we then modify the coregionalisation kernel structure to introduce an additional function output which will be used to model the synthetic fidelity. As shown in section 3, a sample from this expanded coregionalisation GP can be expressed in terms of a linear combination of the preexisting fidelity data and a sample from the intra-fidelity prior. We can then solve for the coefficients that define this linear combination in order to achieve a sample with desired correlation to the ground truth data and any provided lower fidelity proxies.

\begin{figure}[ht]
    \centering
    \includegraphics[height=4.75cm]{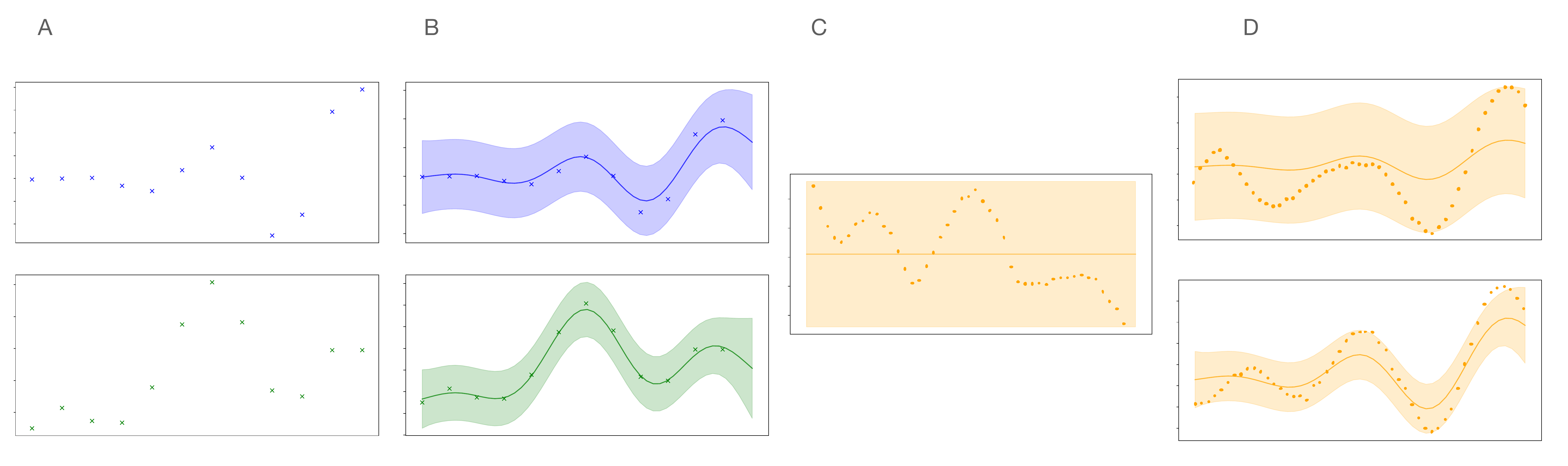}
    \caption{Panels showing progress of algorithm. Panel A shows initial data available for a ground truth function of interest (top blue) and a lower fidelity proxy (bottom green). Panel B shows the data along side the posterior of a MOGP using a coregionalisation kernel that has been fit via maximum log likelihood to the initial data. Panel C shows the prior that uses the fit hyperparameters for the new synthetic fidelity and a sample of the synthetic fidelity drawn from that prior. While Panel D shows two different transformations of the samples and the posterior, chosen to achieve two different correlations to the existing ground truth and lower fidelity proxy.}
    \label{fig:synth_steps}
\end{figure}

A graphical illustration of the steps associated with this algorithm can be seen in figure \ref{fig:synth_steps}

\newpage
\section{Mathematical Background}

% Setup for synthetic fidelity kernel with N tasks where N = 1+J and J is the number of preexisting lower fidelity functions.

% Approach taken to generation of synthetic fidelities 

% Derivation of Pearson correlation of synthetic fidelity posterior mean with high fidelity reference data.

\subsection{Gaussian Process Regression}\label{subsection:GPR}

Gaussian processes express a Gaussian prior over function space where the prior mean is usually taken to be zero and the prior covariance is specified by a  kernel function. Given an initial set of training points represented by an $n$ by $d$ matrix $X$ and associated $n$ dimensional target vector $\vec{y}$, where $n$ is the number of data points and $d$ is the dimensionality of the data, the posterior mean evaluated at a set of test points represented by an $n'$ by $d$ matrix $x_*$ is defined by:

$$\vec{\mu_{\ast}} = K^T_{\ast} K^{-1} \vec{y} $$

where $^T$ indicates the transpose operation, $\vec{\mu_*}$ is an $n'$ dimensional vector of predicted means for the function values at positions $x_*$, $K_*$ is the $n$ by $n'$ matrix defining the covariance between $X$ and $x_*$ and $K^{-1}$ is the $n$ by $n$ inverse covariance matrix between $X$ and $X$. The $K_{\ast}$ and $K$ matrices are computed using the kernel function i.e. $K_{ij} = k_f(x_i, x_j, \theta)$ where $k_f$ is a kernel function, $x_i$ and $x_j$ are the $i$th and $j$th training points respectively and $\theta$ are a set of kernel hyperparameters.

The posterior covariance evaluated at points $x_*$ is defined by:

$$\sigma_{\ast} = K_{\ast\ast} - K^T_{\ast} K^{-1} K_{\ast},$$

where $\sigma_{\ast}$ is an $n'$ by $n'$ posterior covariance matrix and $K_{\ast\ast}$ is the $n'$ by $n'$ prior covariance matrix between $x_*$ and $x_*$.

Once the posterior is calculated we can sample from this posterior via transformation of a white noise vector $\vec{r}$, where the sample over positions $x_*$ is:

$$\vec{s} = \vec{\mu_*} + \sigma_* \vec{r}$$
where $\vec{s}$ is a $n'$ dimensional vector of the posterior sample values and $\vec{r}$ is an $n'$ dimensional Gaussian random vector sampled from $N(0, \mathbb{1})$

We are interested in an MOGP I.e. rather than expressing a distribution over a single set of function values, we express a joint distribution over multiple sets of functional values. To achieve this we augment the domain with an additional dimension that encodes the function index. This allows us to express the covariance between values of different functions evaluated at different points within the domain as a function of points in this augmented domain. We make use of a coregionalisation kernel to calculate this covariance.

In this picture the kernel function is equal to a pairwise value which expresses the relatedness of the fidelities for the two points in question and a standard (typically stationery) GP kernel function. I.e.

$$k^{coreg}_f(x^k_i, x^l_j) = k_f(x_i, x_j) t_{kl}$$

where $x^k_i$ is the $i$th point at fidelity $k$, $x^l_j$ is the $j$th point at fidelity $l$, $k_f(x_i, x_j)$ is a standard GP kernel function and $t_{kl}$ is a coefficient that specifies the covariance between fidelity $k$ and $l$. 

If the training data consists of the value of each fidelity on the same set of training points $X$, the full covariance matrix $K$ can be expressed as:

\begin{displaymath}
    K^{coreg} = \Sigma_T \otimes K_c
\end{displaymath}

where $\Sigma_T$ is an $n_t$ by $n_t$ task matrix made up of the $t_{kl}$ coefficients that encodes pairwise covariances between the $n_t$ different fidelities and $K_c$ is the core task independent kernel matrix between $X$ and $X$.

The values for the $t_{kl}$ are typically fit by maximising the log marginal likelihood of the training data along with any parameters of the kernel function.

\newpage

\subsection{Generating a Posterior over a synthetic fidelity}

We consider the situation where we have a set of points from a domain represented as a $n_x$ by $n_d$ sized matrix $X$ where $n_x$ is the number of points in the domain for which values are available for the ground truth function and any available lower fidelity proxies and $n_d$ is the dimensionality of the domain. The target values are initially represented by an $n_x * n_t$ target vector $\vec{y}$, which is rearranged to form a $n_x$ by $n_t$ matrix $Y$ where $n_t$ is the number of fidelities for which data is available (and is equal to 1 if we only have data from a single reference ground truth function). 

We suppose we have fit the hyperparameters for the prior for a coregionalisation kernel GP via maximisation of the log marginal likelihood and as such have extracted parameters $\theta$ for the core kernel function along with the values of the task matrix $\Sigma_T$. We now wish to consider the case where we will calculate the posterior over an additional task for which there is no data. This will be the synthetic fidelity which we will ultimately sample from. To achieve this we will expand the task matrix to include an additional task. This gives rise to an $n_t+1$ by $n_t+1$ dimensional task matrix $\Sigma'_T$, which requires provision of an additional $n_t+1$ parameters which we initially consider unknown. We are interested in calculating the posterior distribution for this new synthetic fidelity at the same domain points for which the values of the known fidelities are available. I.e. $x_*=X$. We note that this is an extension of the situation examined by Bonilla \cite{bonilla2008multi}.

Inserting the expression for the coregionalisation kernel into the expression for the posterior mean and setting $x_*$ equal to $X$ gives:
%$$\vec{\mu}* = (\Sigma_T \otimes kf*)$$
% $$\vec{\mu_p}=((\Sigma_{T_{n}} \otimes K)(\Sigma_{T_{:n:n}}\otimes K)^{-1}.y$$
% which through rearranged and cancellation gives:

\begin{align*}
\vec{\mu}_p^T &= (\Sigma_{T*}^T \otimes K_c^T)(\Sigma_T \otimes K_c)^{-1} \vec{y}\\
             &= (\vec{\Sigma}_{T*}^T\Sigma^{-1}_{T})Y^T
\end{align*}

% \begin{align*}
% \vec{\tilde{s}} &= \vec{s} - E[\vec{s}]\\
%  &= \sum_{i=0}^{i<n_{t}+1}{c_i\vec{y}_i - \sum_{i=0}^{i<n_{t}+1}{c_iE[\vec{y}_i]}})\\
%  &= \sum_{i=0}^{i<n_{t}+1}{c_i(\vec{y}_i- E[\vec{y}_i]})\\
%  &= \sum_{i=0}^{i<n_{t}+1}{c_i\vec{\tilde{y}}_i}
% \end{align*}

% $$\vec{\mu}_p^T = (\Sigma_{T*}^T \otimes K_c^T) . (\Sigma_T \otimes K_c)^{-1}. \vec{y}$$
% $$\vec{\mu}_p^T=(\vec{\Sigma}_{T*}^T\Sigma^{-1}_{T})Y^T$$
where:
\begin{itemize}
    \item $\vec{\mu}_p$ is the predicted mean values for the synthetic fidelity task,
    \item we have wrapped the target vector $\vec{y}$ into a target matrix $Y$ such that the $i$th column of $Y$ is the values of fidelity $i$ at domain points $x_*$,
    \item $\vec{\Sigma}_{T*}$ is the $n_t$ sized vector of task covariances between the synthetic fidelity and the given fidelity data,
    \item $\Sigma_{T}$ is the $n_t$ by $n_t$ matrix of task covariances associated with the given fidelity data.
\end{itemize}

The equivalent expression for the posterior covariance is:

\begin{align*}
\sigma_p &=(\Sigma_{T**}\otimes K_{c}) -(\vec{\Sigma}_{T*}^T \otimes K_c^T)(\Sigma_T \otimes K_c)^{-1}(\vec{\Sigma}_{T*} \otimes K_c)\\
         &= (\Sigma_{T**}-\vec{\Sigma}_{T*}^T\Sigma^{-1}_{T}\vec{\Sigma}_{T*})K_c
\end{align*}

% $$\sigma_p=(\Sigma_{T**}\otimes K_{c}) -(\vec{\Sigma}_{T*}^T \otimes K_c^T). (\Sigma_T \otimes K_c)^{-1} . (\vec{\Sigma}_{T*} \otimes K_c)$$
% $$\sigma_p=(\Sigma_{T**}-\vec{\Sigma}_{T*}^T\Sigma^{-1}_{T}\vec{\Sigma}_{T*})K_c$$

where $\Sigma_{T**}$ is a scalar defining the task variance associated with the synthetic fidelity

We note the similarity of the above equations to the standard GP equations given in section \ref{subsection:GPR}. The expression for the predictive mean is composed of the $n_t$ sized decorrelated contribution vector of the preexisting fidelities to the synthetic fidelity $\vec{\Sigma}_{T*}^T\Sigma^{-1}_{T}$, which is multiplied by the target matrix $Y$. Thus it is as if we have a Gaussian Process expressed over tasks where rather than scalar valued output we have vector valued output. 

Similarly the expression for the posterior variance strongly resembles the posterior variance of a single fidelity GP where again we replace the prior variance and inverse prior variances over the points in the domain with prior variance and inverse prior variance over the tasks. We can interpret the posterior task variance $\Sigma_{T**}-\vec{\Sigma}_{T*}^T\Sigma^{-1}_{T}\vec{\Sigma}_{T*}$ as the variance that remains after observing the training (task) data (which can be zero if the fidelities for which we have data are sufficiently correlated with the synthetic fidelity). As we are expressing the posterior over a single task - the synthetic fidelity's posterior task variance is a scalar. The full posterior covariance is then equal to this task variance scalar multiple by the prior covariance.

\newpage
\subsection{Generating a synthetic fidelity sample}

A sample of the synthetic fidelity from this posterior is thus:
\begin{align*}
\vec{s} &= \vec{\mu}_p + \vec{\sigma}_p\vec{r}\\
&= Y(\Sigma^{-1}_{T}\vec{\Sigma}_{T*}) + (\Sigma_{T**} - \vec{\Sigma}_{T*}^T\Sigma^{-1}_{T}\vec{\Sigma}_{T*})K_c\vec{r}
\end{align*}

where as before $\vec{r}$ is an $n_x$ sized vector of white noise values.

Examining this expression we can see that a synthetic fidelity posterior sample corresponds to a linear combination of basis vectors $\vec{y_i}$ and sample coefficients $c_i$
$$\vec{s} = \sum_{i=0}^{i<n_{t}+1}{c_i \vec{y}_{i}}$$

where $\vec{y}_{i}$ is column $i$ of $Y$ and $c_i$ is the $i$th component of the vector $(\Sigma^{-1}_{T}\vec{\Sigma}_{T*})$ for $i<n_{t}$, and we additionally define $\vec{y}_{n_t}=K_c\vec{r}$ and $c_{n_t}=(\Sigma_{T**} - \vec{\Sigma}_{T*}^T\Sigma^{-1}_{T}\vec{\Sigma}_{T*})$ to account for the posterior variance.

This can be expressed in terms of an expanded $n_x$ by $n_t+1$ target matrix  $Y'$ as:

$$\vec{s} = Y'\vec{c}$$

where $\vec{c}$ is the vector of sample coefficients $\vec{c}_i = c_i$.

The only dependence of the posterior sample on the task matrix is through the coefficients $c_i$ while the basis vectors $\vec{y}_i$ are independent of the task matrix and as such they can be calculated ahead of time regardless of the desired sample correlation.

\subsection{Specifying a sample with a specific covariance}
Assuming the values of the task matrix $\Sigma_{T}$ have been provided via maximisation of the log-marginal  likelihood, the sample coefficients $c_i$ are then specified by the inter-task covariance between the synthetic fidelity and the given fidelities $\vec{\Sigma}_{T*}$. Typically $\vec{\Sigma}_{T*}$ would be specified by maximisation of the log-marginal likelihood or sampled from an underlying prior; however, in our case we lack data for this synthetic task and more importantly we wish to generate samples from an MOGP posterior such that those samples possess a particular correlation with the preexisting fidelity data. To achieve this we will directly control the values of the coefficients $c_i$ which will implicitly set inter-task covariances $\vec{\Sigma}_{T*}$ such that the generated sample possess the intended correlation. By transforming a set of intended correlations between the synthetic fidelity sample and the existing fidelity data into a set of covariances between the synthetic fidelity sample and the existing fidelity data and augmenting this with a covariance between the synthetic fidelity sample and the sample from the task-independent prior we can formulate a linear problem whose solution will define the sample coefficients we require.

Thus we suppose we have generated the $\vec{y}_i$ basis vectors and have been provided with a vector of Pearson Correlations $\vec{P}_c$ specifying the degree to which the sample should be correlated to the existing $n_t+1$ sample basis vectors, i.e. to the preexisting fidelities and to the sample from the intra-fidelity kernel Prior. (note by the required properties of the full Correlation Matrix, specification of the correlation between the synthetic fidelity sample and the task independent prior sample is determined given specification of the correlations to the preexisting fidelities).

By making use of a heuristic defined below we choose a variance $\vec{\sigma}_h$ for our intended sample and use this to transform the vector of Pearson Correlations into a vector of sample covariances $\vec{\sigma}_s$:

$$\vec{\sigma}_s = \Sigma^{0.5}_h \cdot \vec{P}_c$$

where $\Sigma_h^{0.5}$ is a diagonal matrix with diagonal elements $\sqrt{\sigma_h}$.

\newpage
We define zero mean sample basis vectors as:

$$\vec{\tilde{y}}_i = \vec{y}_i - E[\vec{y}_i].$$

Similarly we define the zero mean sample as:
\begin{align*}
\vec{\tilde{s}} &= \vec{s} - E[\vec{s}]\\
 &= \sum_{i=0}^{i<n_{t}+1}{c_i\vec{y}_i - \sum_{i=0}^{i<n_{t}+1}{c_iE[\vec{y}_i]}})\\
 &= \sum_{i=0}^{i<n_{t}+1}{c_i(\vec{y}_i- E[\vec{y}_i]})\\
 &= \sum_{i=0}^{i<n_{t}+1}{c_i\vec{\tilde{y}}_i}
\end{align*}

i.e.
$$\vec{\tilde{s}} = \tilde{Y'}\vec{c}$$

where $\tilde{Y'}$ is the matrix of normalised sample basis vectors $\tilde{Y'}_i = \vec{\tilde{y}}_i$.

As we have transformed the Pearson Correlations into covariances and the dot product between two zero mean sample vectors gives the covariance, we can express the requirement that the covariance between the synthetic fidelity sample and the sample from fidelity $i$ be equal to $\sigma_{si}$ as:

$$\vec{\tilde{s}}\cdot\vec{\tilde{y}}_i = \sigma{_{s_i}}$$

i.e. 
$$\tilde{Y'}^T\vec{\tilde{s}}=\vec{\sigma}_s.$$

By substituting the expression for $\vec{\tilde{s}}$ above we arrive at:

$$\tilde{Y'}^T\tilde{Y'} \vec{c} = \vec{\sigma}_s$$

which we can solve for the basis coefficients:

$$\vec{c} = (\tilde{Y'}^T\tilde{Y'})^{-1} \vec{\sigma}_s$$

and hence generate the required sample.

\newpage
\subsection{Heuristic for transformation of sample Pearson Correlations to sample covariances}

It is natural to specify desired correlations to a set of preexisting fidelity data; however, to solve for the desired sample we require transformation of the correlations into a set of covariances. This can be achieved by specification of a desired sample variance. We make use of a heuristic to pick this sample variance that ensures that a correlation vector that specifies 100\% correlation to one of the preexisting fidelities will be assigned a variance equal to the variance of the matching fidelity.

The proposed heuristic is to use the overlap of the vector $\vec{P}_c$ with the rows of the $n_{t}+1$ by $n_{t}+1$  correlation matrix $C$ associated with the sample basis vectors defined as.

\begin{align*}
C_{ij} = \Sigma_i^{-0.5}\vec{\tilde{y}}_i \cdot \Sigma_j^{-0.5}\vec{\tilde{y}}_j
\end{align*}

where $\Sigma_i$ is a diagonal matrix with diagonal elements equal to the variance of $\vec{\tilde{y}}_i$.

 We then iteratively determine a set of overlap coefficients by picking the maximally overlapping correlation matrix row vector, noting the overlap then projecting out the selected vector from $P_c$ and continuing until we have a set of overlaps associated with each sample basis. I.e.

\begin{align*}
&\vec{v} \leftarrow \vec{\mathbb{0}}\\
&\vec{w} \leftarrow \vec{\mathbb{0}}\\
&\text{for} \; j < n_t +1:\\
& \;\;\;\vec{P}_c \leftarrow \vec{P}_c - (\vec{P}_c \cdot \vec{v})\vec{v} &\\
& \;\;\;i \leftarrow \text{arg\,max}_{i}(\vec{C}_i \cdot \vec{P}_c) &\\
& \;\;\;\vec{v} \leftarrow \vec{C}_i &\\
& \;\;\;w_j \leftarrow \vec{C}_i \cdot \vec{P}_c &\\
\end{align*}

We then determine the sample variance by computing the weighted mean of the sample basis variances. I.e.
$$\vec{\sigma}_h = \vec{w} \cdot \text{diag}(\tilde{Y}^T\tilde{Y})$$

\subsection{Generating valid Pearson Correlation Vectors}

The above assumes provision of a vector of desired Pearson Correlations $\vec{P}_c$ to the $n_t+1$ prior fidelities and to the sample from the prior. The constraints on such a vector are well known; however, for completeness we outline the generation procedure. 

We suppose that the elements of the $\vec{P}_c$ vector are chosen sequentially and that the bounds for the values of a particular element of the $\vec{P}_c$ vector are conditioned on the previous choices. I.e we start with a free choice of an initial correlation value of the synthetic fidelity sample to one of the $n_t$ fidelities with bounds of $[-1, 1]$, then subsequent choices are constrained such that the combined correlation matrix as defined above is a valid correlation matrix. I.e. is positive semi-definite with diagonal elements equal to one.

Given the set of sample basis vectors we calculate the correlation between them, giving an $n_{t}+1$ by $n_{t}+1$ correlation matrix $C$. Then via Cholesky decomposition we have:

$$C = U^T U$$

where $U$ is an upper triangular matrix. We now consider the expansion of $C$ by an additional row and column to form an $n_{t}+2$  by $n_{t}+2$ combined correlation matrix $C'$ where this additional row/column will be the Pearson correlation vector $\vec{P}_c$. The expansion of $C$ to $C'$ leads to an equivalent expansion of the Cholesky factorised matrix $U$ to $U'$. 

Via the Cholesky decomposition, for each $i$ we have 

$$\vec{U}'_i \cdot \vec{U}'_{p} = P_{ci}$$

where $\vec{U}'_i$ is the $i$th column of $U'$ and, for ease of notation, we let $p=n_t+1$ be the index of the last column.

Since $U'$ is an upper triangular matrix it follows that:

$$\vec{U}'_i \cdot \vec{U}'_{p} = \vec{U}'_{[:i],i} \cdot \vec{U}'_{[:i],p} + U'_{i i} U'_{i p}$$
where $\vec{U}'_{[:i],j}$ is the vector corresponding to the first $i$ elements of column $j$ of the Cholesky factor matrix $U'$.

We can use this relation to provide bounds on the possible values of $\vec{P}_c$, and we do so sequentially by providing bounds on $P_{ci}$ in terms of the previous values $P_{c0} \ldots P_{c,{i-1}}$.

As the Cholesky factor for a correlation matrix has element $U'_{00}=1$ for initial choice of correlation $P_{c0}$ we have: 

$$U'_{0 p} = P_{c0}$$ 

Further, as the correlation of the synthetic fidelity sample with itself must be 1 it follows that $\vec{U}'_{p}\cdot \vec{U}'_{p} = 1$. Then, since

$$\vec{U}'_{p} \cdot \vec{U}'_{p} = \vec{U}'_{[:i],p} \cdot \vec{U}'_{[:i],p} + {U'_{i p}}^2 + \sum_{j=i+1}^p {U'_{j p}}^2$$

this means that the bounds on the allowed values for $U'_{i p}$ can be expressed in terms of the previous values $\vec{U}'{_{[:i],p}}$ as follows:

$$\vec{U}'_{[:i],p} \cdot \vec{U}'_{[:i],p} + {U'_{i p}}^2 \leq 1$$

i.e.

$$ -\sqrt{1-\vec{U}'_{[:i],p} \cdot \vec{U}'_{[:i],p}} \leq U'_{ip} \leq +\sqrt{1-\vec{U}'_{[:i],p} \cdot \vec{U}'_{[:i],p}} $$

Which translates into bounds for the Pearson correlation coefficient $P_{c i}$ of:

$$\vec{U}'_{[:i],i}\cdot \vec{U}'_{[:i],p} - U'_{ii}\sqrt{1-\vec{U}'_{[:i],p} \cdot \vec{U}'_{[:i],p}}  \leq P_{c i} \leq \vec{U}'_{[:i],i}\cdot \vec{U}'_{[:i],p} + U'_{ii}\sqrt{1-\vec{U}'_{[:i],p} \cdot \vec{U}'_{[:i],p}} $$

\newpage
\section{Pseudo-Code}
\begin{lstlisting}
def draw_s(K, X, Y):
    # generate sample from the prior
    r = normal(0, ones_like(X))
    y_n = (K @ r).t()
    
    # set prior sample std to mean of Y std to make heuristics choice 
    # of sample std more intuitive 
    y_n /= y_n.std()
    y_n *= Y.std(axis=1).mean()
    
    # create set of reference vectors
    Y_ = vstack([Y, y_n])
    Y_ -= Y_.mean(axis=1)[:,None]
    
    # calculate reference covariances
    fid_cov = (Y_@Y_.t())
    fid_stds = diag(fid_cov)**0.5
    
    # calculate reference Pearson Correlations
    _Y_ = Y_/fid_stds[:, None] 
    fid_pcs = _Y_ @ _Y_.t()
    pcs = build_pcs(fid_pcs)

    heuristic_weights = get_heuristic_weights(pcs, fid_pcs)
    synth_fid_std = heuristic_weights @ fid_stds

    # using the reference vector stds and the calculated synth_fid std 
    # generate a set of covariances from the supplied pearson correlations
    sigma_s = pcs * fid_stds * synth_fid_std
    
    # using the covariances solve for the coefficients
    c = inverse(fid_cov) @ sigma_s
    
    # use the coefficients to generate a sample
    s = c[:,None].t() @ Y_
    
    return s
\end{lstlisting}

\newpage

\begin{lstlisting}
def get_heuristic_weights(pcs, ref_C):
    weights = []
    for i in range(len(ref_C)):
        overlaps = ref_C @ pcs
        ind = argmax(overlaps)
        weight = overlaps[ind]
        weights.append(weight)
        pcs = pcs - weight*ref_C[ind]
        
    return array(weights)    

def build_pcs(ref_C):
    """Sequentially build up Pearson correlation values for synthetic sample 
    against known fidelities and prior sample"""
    ref_L = cholesky(ref_C)
    
    # expand cholesky matrix with additional row/column
    master_L = zeros([len(ref_C)+1, len(ref_C)+1])
    master_L[:-1, :-1] = ref_L
    master_L[-1, -1] = 1
    
    pcs = []
    for i in range(len(ref_C)+1):
        partial_pc_i = (master_L[-1,:i] @ master_L[i,:i])
    
        abs_max_l_i = (1 - master_L[-1,:i]@master_L[-1,:i])**0.5
        
        if i != len(ref_C):
            bound_pc_i =  (master_L[i,i]*abs_max_l_i)
        else:
             bound_pc_i =  (abs_max_l_i**2)
                            
        upper = partial_pc_i + bound_pc_i
        lower = partial_pc_i - bound_pc_i
                       
        allowed_vals = [lower, upper]
        pc = get_input(allowed_vals)
        pcs.append(pc)
                       
        # convert pc back in l and add to the master_L matrix
        _l = (pc - partial_pc_i)/ master_L[i,i]
        master_L[-1,i] = _l
        
    return pcs
    
\end{lstlisting}    

\newpage
\section{Examples}

In this section we graphically show examples generated for the Liu \cite{Liu2018b} and Currin \cite{currin1991bayesian} synthetic functions making use of the ground truth and low fidelity proxy functions within a dual task coregionalisation kernel MOGP which in turn made use of a spectral mixture model \cite{Parra2017} with 4 mixtures as the intra-fidelity kernel.

\begin{figure}[h]
    \centering
    \includegraphics[height=5.5cm]{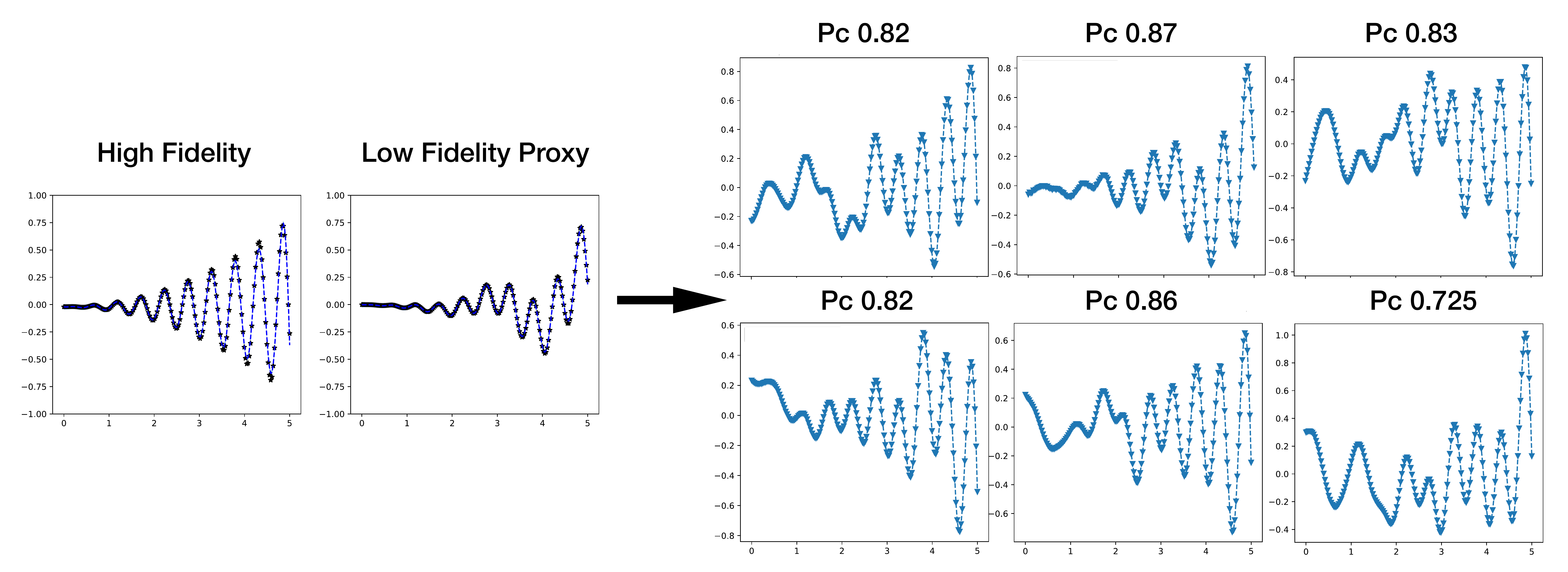}
    \caption{The left panel shows reference high and and low fidelity function data generated using Liu's functions, where the data points used for training are shown by the markers and the underlying functions are shown by the dashed lines. The right panel shows a selection of six synthetic fidelity samples that have been generated with varying sample correlations to the high fidelity, predicted values are again shown with markers while the dashed line shows an interpolation of the data}
    \label{fig:eg_liu}
\end{figure}

\begin{figure}[h]
    \centering
    \includegraphics[height=5.5cm]{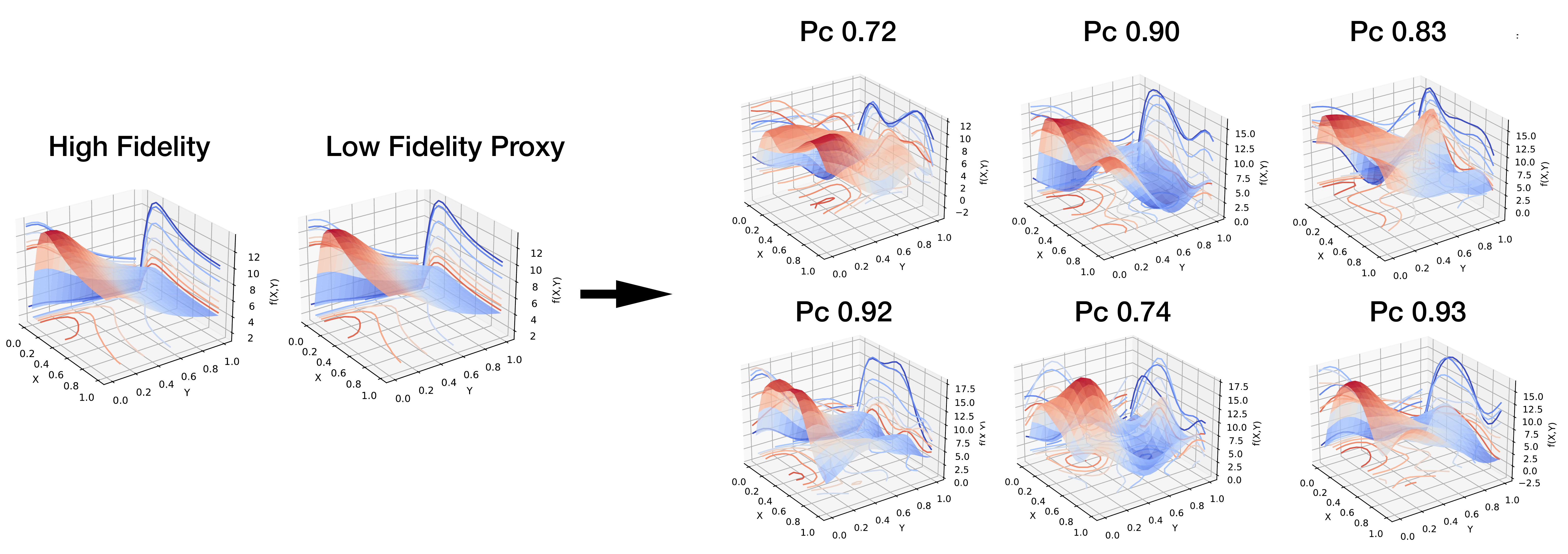}
    \caption{The left panel shows reference high and low fidelity function data generated using Currin's functions. Data points were taken from a 20 x 20 grid. To aid visualisation a contour plot of the function values is present on the x-y plane in addition to  projections of the function values onto the x-f and y-f planes. Colouring of the function surface and x-y contour plot reflect the value of the function while colouring of the function projections reflects the distance of the function value from the projection plane.The right hand panel shows a selection of six synthetic fidelity samples that have been generated with varying sample correlation to the high fidelity.}
    \label{fig:eg_currin}
\end{figure}

\newpage
\section{Conclusion}

In this work we present an initial approach to the problem of systematic synthesis of datasets for the evaluation of multi-fidelity optimisation problems. We show that we can construct a multi-output Gaussian process making use of the coregionalisation kernel based on high fidelity data from a ground truth function and any lower fidelity proxies that are available and expand this GP it to include a synthetic fidelity. We show that samples drawn from the posterior of this synthetic fidelity can be expressed in terms of a linear combination of the known fidelity data and a sample from the prior and we use this to generate synthetic samples that are arbitrarily correlated to the ground truth function and the available lower fidelity proxies. We anticipate this approach will allow more detailed interrogation into the factors that influence multifidelity optimisations in particular the relationship between lower fidelity proxy costs and interfidelity correlation on regret curves.

\section{Acknowledgement}

This work was supported by the Hartree National Centre for Digital Innovation, a collaboration between STFC and IBM.
% IBM and HNCDI

% \begin{acks}
% \end{acks}

\bibliographystyle{ACM-Reference-Format}
\newpage
\bibliography{SyntheticFidelity.bib} 

\end{document}